# Human-like Driving Decision at Unsignalized Intersections Based on Game Theory

Daofei Li[1], Guanming Liu[1], Bin Xiao[1]

**Abstract:** Unsignalized intersection driving is challenging for automated vehicles. For safe and efficient performances, the diverse and dynamic behaviors of interacting vehicles should be considered. Based on a game-theoretic framework, a human-like payoff design methodology is proposed for the automated decision at unsignalized intersections. Prospect Theory is introduced to map the objective collision risk to the subjective driver payoffs, and the driving style can be quantified as a tradeoff between safety and speed. To account for the dynamics of interaction, a probabilistic model is further introduced to describe the acceleration tendency of drivers. Simulation results show that the proposed decision algorithm can describe the dynamic process of two-vehicle interaction in limit cases. Statistics of uniformly-sampled cases simulation indicate that the success rate of safe interaction reaches 98%, while the speed efficiency can also be guaranteed. The proposed approach is further applied and validated in four-vehicle interaction scenarios at a four-arm intersection.

## 1. Introduction

Intersection driving accounts for large amounts of accidents in both urban and rural areas. A query of Fatality Analysis Reporting System (FARS) shows that from 2015 to 2019, there were 169,009 fatal crashes across the United States, while 24.7% of them occurred at intersections.[1] Comparing with that in signalized intersections, the situation gets much more challenging when driving in unsignalized intersections. Liu et al. shows that the accident rate and severity at unsignalized intersections in China are both higher than that at signalized intersections.[2] By nature, intersection driving requires more dynamic interactions with other road users, including perception, prediction of intent and future motion, negotiation/cooperation and decision, etc.

Unsignalized intersections are extremely challenging for automated driving agents, too. Comparing with human drivers, the current technology of automated intersection decision algorithms is still clumsy. Safety always comes first, but an over-conservative driving agent hesitating at the intersection means low traffic efficiency, unacceptable user experience, and confusing the drivers of other interaction vehicles. Therefore, it is crucial to find a safe and efficient driving strategy for unsignalized intersections.

### 1.1 Literature review

1.1.1 Centralized methodologies

A variety of algorithms for intersection decision have been proposed in recent years, which can be divided into two categories according to different research perspectives. Category I algorithms adopt centralized methodologies, and focus on the overall efficiency of the transport system. Colombo and Del Vecchio designed a controller that can effectively solve the collision avoidance of multiple vehicles at intersections under the framework of a scheduling problem.[3] Pourmehrab et al. proposed a system design for both automated/connected and conventional vehicles approaching an isolated intersection, which can maintain shorter and more stable headways to the leading vehicles.[4] Pei et al adopted sequential decomposition to decompose the large-scale planning of road networks into distributed sub-problems of finding cooperative driving strategy. The overall safety and efficiency of the road networks are ensured by assigning the arrival time of each vehicle entering the intersection along the designed route.[5] However, these algorithms usually require vehicle connectivity and advanced resources of centralized computing. Furthermore, such centralized methodologies can not consider the specific interactions between individual drivers.

1.1.2 Single-vehicle decision methodologies

On the other hand, Category II studies pay more attention to the single-vehicle decision. There are basically two ways to treat the decision modelling, i.e. learning-based and mechanism-based approaches.

**Learning-based approaches**

Learning-based approaches are basically data-driven, and the popular learning frameworks include deep learning and reinforcement learning. In the training process, traffic situation variables and final decisions work as inputs and outputs of model, respectively. Tram et al. (2018) considered the interaction between automated and human-driven vehicles, and proposed a decision algorithm by learning typical behaviors of other vehicles.[6] Tram et al. (2019) further suggested another decision algorithm based on reinforcement learning, which can learn from the yielding, following and overtaking behaviors of drivers.[7] Taking the unprotected left-turn decision as an example, Oh et al. utilized support vector machine to classify the yield-or-not decision made by two kinds of human drivers.[8] However, the trained hyperplanes, as the decision criteria, vary greatly with different driving styles, which calls for an additional module of driving style recognition in potential applications. As the name suggests, the learning-based approaches is certainly data-hungry, which means that an effective decision model should be based on high-quality training data.

**Mechanism-based approaches: game-theoretic algorithms**

Mechanism-based approaches focus on the key principles working in a driving decision, i.e. how to map from stimuli (traffic situation variables) to responses (final decisions). Like interpersonal communication, the decision-making in interactive driving needs clear understanding of each other, especially the driving decision style. Among other frameworks, game theory has been adopted to model the intersection driving behaviors in many studies.[9–14] Mandiau et al. used the classic game theory to analyze the driver coordination mechanism, while there were only limited details about the payoff design in decision matrices.[9] They also presented an approach to simplify multi-vehicle interaction problems to a series of two-vehicle interaction sub-problems. Liu and Wang regarded the vehicle-crossing process as a dynamic repeated game, and built a driver model to describe two types of driving, i.e. cautious and risky.[10] Recently, Tian et al. proposed an approach based on the level-k game theory to model vehicle interactions at urban unsignalized intersections, and imitation learning was used to obtain the decision policies.[11] Li et al. presented a Leader-Follower game-theoretic framework for various parametrized intersection scenarios.[12] Interacting vehicles were assigned into the roles of leader or follower based on the right of way, and then different decision-making strategies for the leader and the follower were designed accordingly. Receding horizon optimization was successfully introduced for acceleration decision, but a fixed prediction horizon might not work in the intersection scenarios. Similarly to the work of Mandiau et al, they also regarded a multi-vehicle game as a series of two-vehicle games. Wang et al. proved the existence of Nash equilibrium in both competitive and cooperative games in the proposed unsignalized intersection model.[13] They utilized the branch and bound (B&B) paradigm to solve the optimal action sequence during the entire time horizon of passing the intersections. More recently, Rahmati et al. formulated the scenario of unprotected left turn as a Stackelberg game, for which the field data were used to calibrate the parameters in the payoff functions.[14] To summarize, game theory has been proved promising in the intersection decision problem, both in theoretical and practical senses. The earlier models focus on the qualitative analysis of feasibility, especially how a formulated game can handle intersection interactions with two or more participants. In contrast, the recent researches in this field start to focus more on the detailed game formulation for specific scenarios. For real applications, there are still many issues requiring further study, e.g. flexible game formulations and effective payoff designs that can handle complex interactions.

On the other hand, the driving style of interacting human drivers has been considered in some researches. Kaysi and Abbany built a behavioral model to predict the other drivers' probability to drive aggressively at unsignalized intersections.[15] Taking unsignalized T-junction intersection as an example, Sezer et al. designed a collision avoidance model based on Mixed Observability Markov Decision Process, which can predict the temperament and intention of the interacting driver before the yield decision.[16] Pruekprasert et al. used sequential games to make decision by considering four kinds of interacting driver styles, and they found that the algorithm can perform better if considering irrational vehicles.[17] However, in these existing studies the interacting driver behaviors are just classified into a few types, which is not sufficient for real applications and calls for a more flexible framework to describe the rapid-changing dynamics of interaction.

1.1.3 Human-likeness in decision

Considering that automated vehicles (AV) are expected to interact with other human drivers in the foreseeable future, any AV that can fit in traffic with manually-driven vehicles should decide with human-likeness. Many existing literatures on automated driving decision have attempted to realize human-like and thus traffic-compatible strategies in driving context, though not all of them use the word "human-likeness" in description. A natural and common way is to directly imitate the human driver decision by learning[6–8]. Beaucorps et al. applied clustering to human driving data in unsignalized intersections and roundabouts, to achieve human-like speed control in complex interactions[18]. Based on the actual human driver data, Gu et al presented a human-like motion planning model to represent how drivers make complicated interactions with pedestrians in signalized intersections, including how to detect pedestrian intention and select a gap between pedestrians to pass[19]. Based on CARLA simulator data, a deep imitation learning framework is proposed to design the driving policy for complex urban scenarios[20]. Li et al proposed an end-to-end decision-making method with pyramid pooling convolutional neural network and long-short-term memory, while both CARLA data and a naturalistic driving dataset were used for model training[21]. Gonzalez et al reported a human-like decision approach using Partially Observable Markov Decision Process (POMDP), which can mimic the human abilities of anticipating surrounding drivers' intentions in highway driving[22]. Seong et al proposed an attention-based deep reinforcement learning framework for interactive driving at unsignalized intersections, which can realize human-likeness by learning to focus on spatially and temporally important features[23]. Another way to realize human-likeness is to incorporate human decision rules, styles or preferences in mechanism-based algorithms. For example, Hang et al recently proposed a human-like decision-making approach using a non-cooperative game theoretic framework, which can represent three different human driving styles in trading-off among safety, comfort and travel efficiency in lane changing scenarios[24].

To sum up, it can be said that (1) human-likeness in decision can be reflected in the modules of predicting other road users' intention and motion, assessing the current and future collision risks, and trading-off among multiple performance indices, etc. (2) Both learning- and mechanism-based approaches can work for human-like driving decision, while learning-based approaches need an appropriate learning framework and also a complete coverage of driving data. Mechanism-based approaches rely more on explicit human decision models in strategic and tactical levels and can be more flexible and scalable in different applications.

As a mechanism-based approach, game theory is a promising framework for the driving decision at unsignalized intersections. However, even with a same general framework of game, a different view about the intersection decision game can lead to different forms and advantages of solution. In highly dynamic interactions, the decisions of both drivers are coupled together, and accordingly the decision styles are dynamically coupled, too. Therefore, there is a research need to further develop an explicit model to describe such dynamic interaction characteristics. Particularly, to achieve human-likeness in balancing safety and speed, a new methodology of mapping the decision factors to payoffs is necessary.

### 1.2 Main contribution

To address these needs, the main contributions of this study are three-fold. (1) To capture the human-likeness in interactive driving decisions, Prospect Theory is introduced to map the objective collision risk to the subjective driver payoffs, and the driving style can be quantified as a tradeoff between safety and speed. (2) A probabilistic model is introduced to describe the driver acceleration tendency in a two-vehicle interaction at unsignalized intersections. (3) To realize such human-like decision capabilities, a game-theoretic framework is proposed and implemented for unsignalized intersection driving, of which the advantages are validated by simulations.

The rest of the paper is organized as follows. Section 2 defines the decision problem at intersections and leads to a simplified problem of two-vehicle interaction. Section 3 introduces the hierarchical algorithm structure and then details the game formulation of decision problem, including objective and subjective payoffs. Section 4 presents the simulation results in limit cases of intersection driving, including two- and four-vehicle interactions. The conclusion is provided in Section 5.

## 2. Problem definition of intersection driving decision

### 2.1 Multi-vehicle conflicts at intersection

First, the key elements of the multi-vehicle conflict problem are defined. The set of vehicle participants at the intersection is defined as $N = \{1,2,3,\ldots,n\}$, and the special participant $N_0$ means the intersection itself. The public information set at the intersection is defined as $I = \cup_{i \in N \cup N_0} I_i$, while $I_i$ means the state information that the vehicle can directly observe, e.g. the speed, acceleration, position, and geometry dimensions of vehicles. The set of possible driving path is $P = \{Left, Right, Straight\}$, representing turning left, turning right and going straight, respectively. The set of strategies for vehicle $i$ is defined as $S_i, \forall i \in N$, and the probability distribution of its intended path is $P_i = [P_{i,left}\ P_{i,right}\ P_{i,straight}], \forall i \in N$. Its payoff function of safety and speed is $R_i, \forall i \in N$. Therefore, the multi-vehicle interaction problem at the intersection $N_0$ can be defined as

$$G_{N_0} \stackrel{\text{def}}{=} \langle N, I, \{S_i\}, \{P_i\}, \{R_i\}\rangle, i \in N \quad (1)$$

Considering that there have been lots of studies on trajectory prediction, it is assumed that the intended path $\{P_i\}$ is determined. So, the considered problem is reduced as $G_{N_0} = \langle N, I, \{S_i\}, \{R_i\}\rangle, i \in N$. In this paper, the problem $G_{N_0}$ is set to be the most basic problem with only two participants, i.e. $N = \{1,2\}$. If with more than two participants, i.e. n>2, the problem of $G(\text{n} > 2)$ can be decomposed into n-1 sub-problems of $G(\text{n} = 2)$, for which the best strategy $s_i^* \in S_i$ can be found from the intersection of optimal sub-problem solution sets, i.e.

$$s_i^*(N = \{1,2 \ldots, n\}) = \cap_{j \in N, j \neq i} s_i^*(N = \{i,j\}) \quad (2)$$

This simplification approach has been proved effective in reducing the computational complexity.[9,12]

Ideally, the intersection of such optimal solution sets is not null, so the best strategy is the one that corresponds to the maximum payoff. If there is no intersection of optimal solution sets in all sub-problems, meaning that there is no optimal strategy satisfying all two-vehicle interaction sub-problems, a most conservative strategy, even emergency braking strategy $s_{AEB}$, is chosen. As a special case, if only some sub-problems have optimal solutions, it is also treated as a case of "no intersection of optimal solution sets in all sub-problems" and thus a conservative strategy is chosen.

For the optimal decision problem with only two vehicle participants, $G_{N_0} = \langle N, I, \{S_i\}, \{R_i\}\rangle, N = \{1,2\}$, there are a total of 9 situations for different path forms in a four-armed intersection. As shown in Fig. 1, to simplify the interaction problem, the shaded rectangular area is the potential conflict area of vehicles. According to whether there are conflict areas in the vehicle driving paths, the 9 interaction situations can be divided into 3 categories. (1) The vehicle paths overlap only at one conflict area, as shown in case (a). (2) The vehicle driving paths do not overlap, as in case (b). (3) The vehicle paths have multiple overlapping areas, as in case (c). The case (c) problem can be transformed to a multi-vehicle interaction problem in case (d), in which the host vehicle interacts with the other two virtual vehicles. Then again it reduces to a problem of two-vehicle interacting with only one conflict area in case (a). Therefore, the two-vehicle interaction problem in case (a) is the core problem to be discussed, while the other intersection decision problems can be regarded as its extended forms.

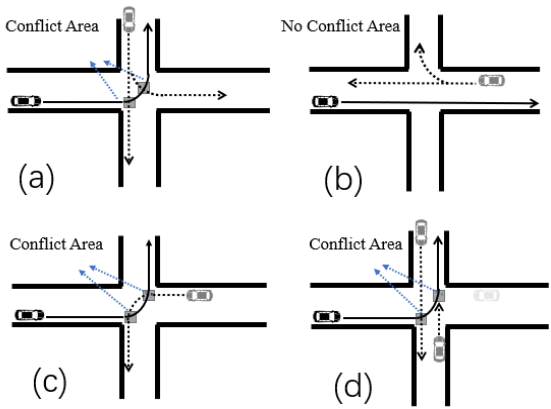

Fig. 1 Two-vehicle interaction in an unsignalized four-armed intersection

### 2.2 A Two-vehicle Example

Fig. 2 schemes an example of two-vehicle interaction driving in an unsignalized intersection. The longitudinal decision "to accelerate" or "to decelerate" will be the focus problem, while the path following in the lateral direction is not covered here. For collision judgement after the entire interaction, residual clearance $\Delta_d$ is defined as the distance of the late-arrival vehicle (Car A) away from the conflict area, when the early-arrival vehicle (Car B) just arrives at the conflict area. For safe intersection crossing, the residual clearance cannot be too small, otherwise it may activate emergency braking of both vehicles and even cause collision.

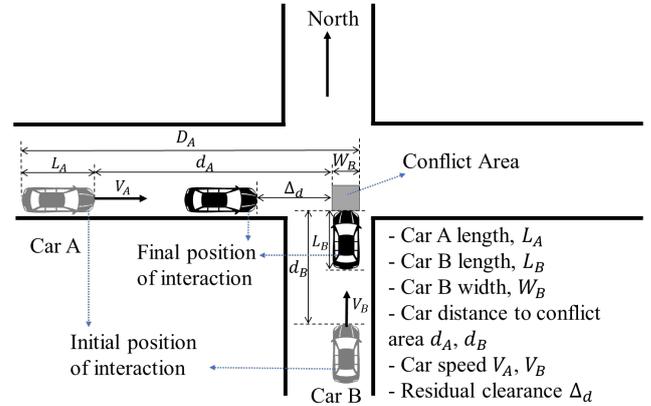

Fig. 2 Two vehicle interaction in an unsignalized intersection

## 3. Game-based formulation for intersection decision

### 3.1 Hierarchical decision structure

During the interaction at unsignalized intersection, it is necessary for the host vehicle to speculate on the intention and behavior of the interaction vehicle, and then to make a complete decision. Hence for flexibility and compatibility when applied in different vehicles, a hierarchical decision structure is adopted, by decoupling the tactical and operational decisions into two levels. As shown in Fig. 3, the top level is to determine "to accelerate" or "to decelerate". Here, maintaining the current speed is considered as a "to accelerate" strategy. In the bottom-level the second decision is to optimally generate a target acceleration $a_x^*$, for which vehicle limitations of acceleration/braking, comfort and emergency braking condition should be considered. Before feeding into the PID controller in tracking, the demand acceleration $a_x^*$ is filtered by a first-order block, $1/(\tau s + 1)$, to avoid abrupt change of acceleration for better comfort. However, a bigger time constant $\tau$ represents that the transient response before steady state lasts longer, which may result in a larger error between $a_x^*$ and $a_{xf}^*$ and thus harm the decision performance. Here, $\tau$ is chosen as 0.5 for the considered electric vehicle. Finally, the PID controller tracks the demand acceleration and outputs throttle and brake pressure to the vehicle dynamics model to update the vehicle states.

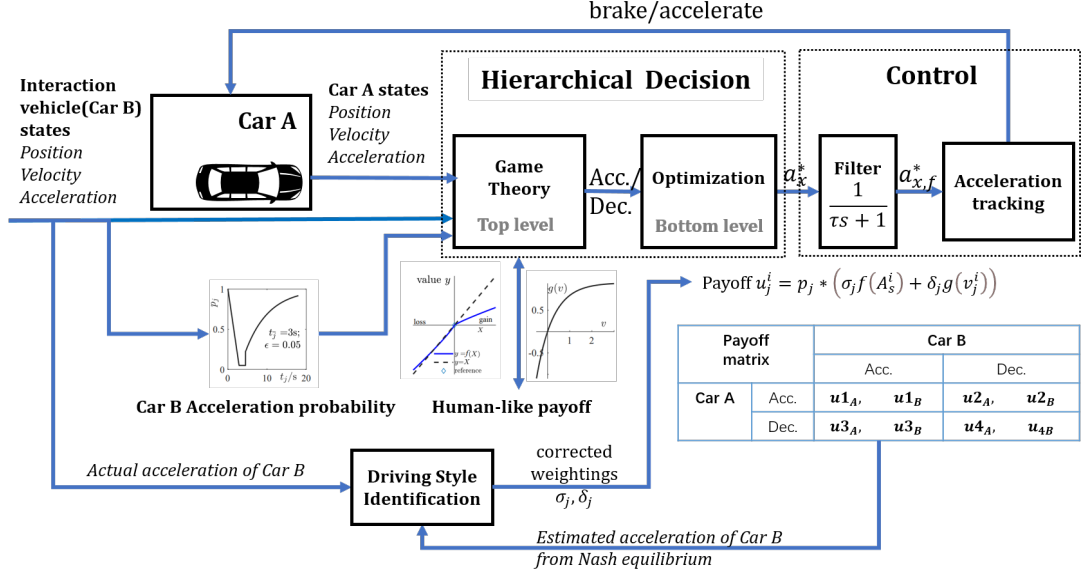

Therefore, the dynamic decision of intersection driving can be formulated as a multi-stage static game process. In every subgame, both players of the game take choices and perform actions at the same time. The gaming elements are as follow:
a) the sub-game duration *T*, also works as the decision-making interval;
b) the participant set {A, B}, indicating Cars A and B, respectively;
c) the information set, including the current speed, acceleration and position of both A and B;
d) the strategy set $\{a_{Acc}, a_{Dec}\}$, where $a_{Acc}$ indicates acceleration and $a_{Dec}$ indicates deceleration;
e) the payoff functions, including safety and speed payoffs.

The potential advantage of game-based formulation of intersection driving depends on the accurate modeling of the interaction between two vehicles. Therefore, the payoff design for two interacting agents, human or non-human, is the essential part of algorithm development.

### 3.2 Objective performance indices
3.2.1 Safety

For objective indicators, it is natural to pick some safety-related and efficiency-related variables for corresponding payoffs. To ensure safety of two vehicles crossing the intersection, the necessary and sufficient condition is that the tail of the first-arrival vehicle has already completely passed the conflict area, when the head of the late-arrival vehicle just arrives at the conflict area. By using kinematics, it's presented how a two-vehicle interaction can be evaluated from the view of safety. For compact expression of variables, *j* in subscripts is used to indicate the subject vehicle, and *i* in superscripts is used to indicate the *i*-th subgame.

For vehicle *j* in the *i*-th subgame, the arrival time $t_j^i$ is set as the time that it needs to arrive at the conflict area (also called time to arrival, TTA) and passing time $\tau_j^i$ is set as the time it needs to completely pass the conflict area, while keeping its current speed and acceleration. The kinematic descriptions of $t_j^i, \tau_j^i$ are as follows:

$$t_j^i = \text{sign}(a_j^i) * \sqrt{(V_j^i/a_j^i)^2 + 2\,d_j^i/a_j^i} - V_j^i/a_j^i, \quad (3)$$

$$D_j^i = d_j^i + L_j + W_{\bar{j}}, \quad (4)$$

$$\tau_j^i = \text{sign}(a_j^i) * \sqrt{(V_j^i/a_j^i)^2 + 2\,D_j^i/a_j^i} - V_j^i/a_j^i, \quad (5)$$

where the subscripts $j$ and $\bar{j}$ refer to the host and interaction vehicles, respectively. $L_j$ is the length of vehicle *j* and $W_{\bar{j}}$ is the width of its interaction vehicle $\bar{j}$. $d_j^i$ is the distance from the head of vehicle *j* to the near edge of the conflict area. $D_j^i$ is the distance from the tail of vehicle *j* to the far edge of the conflict area. $V_j^i$, $a_j^i$ are the current speed and acceleration of vehicle *j*, respectively.

Note that if a vehicle stops at a position far away from the intersection, i.e. $a_j^i = 0$, then arrival time $t_j^i = t_{max}$ and passing time $\tau_j^i = t_{max}$, where $t_{max}$ is a limit value.

If arrival times $t_A^i > t_B^j$, Car B will firstly arrive at the conflict area at current speed and acceleration, so it has the priority to cross the intersection. If $t_A^i = t_B^j$, the vehicle on the right side, i.e. Car B, is supposed to have the priority to cross according to traffic rules. As shown in Fig. 2, corresponding to the residual clearance $\Delta_d$, the estimated residual interval $\Delta t^i$ is defined as the time difference between Car B passing and Car A arriving at the conflict area:

$$\Delta t^i = t_A^i - \tau_B^i \quad (6)$$

The estimated residual interval $\Delta t^i$ is a variable that can represent the safety degree of interaction. If $\Delta t^i \geq 0$, the two vehicles can safely pass the intersection, while if $\Delta t^i < 0$, there will be a collision. As for special cases when one vehicle $j$ stops at a position far away from the intersection, i.e. $t_j^i = t_{max}$, then the maximum value for estimated residual interval $\Delta t^i$ is denoted as $\Delta t_{max}$.

Assuming in next step Cars A and B will adopt the expected strategies $a_{expA}$ and $a_{expB}$, respectively, it's able to calculate the expected residual interval $\Delta t_{exp}^i$ according to Eqs. (3) to (6), as an indicator of safety payoff. To further consider the extra effect of the new strategy on safety, the increment $\Delta t_{exp}^i - \Delta t^i$ is used to represent a correct direction of improvement. Finally, as an objective safety indicator, the safety advantage $A_s^i$ is defined as,

$$A_s^i = \Delta t_{exp}^i + w_t * (\Delta t_{exp}^i - \Delta t^i) \qquad (7)$$

where $w_t$ is a weighting coefficient, here $w_t$=0.5.

3.2.2 Speed

If using an expected acceleration strategy $a_{expj}$, the future speed of vehicle $j$ after a subgame duration $T$ is

$$V_j^{i+1} = V_j^i + T * a_{expj} \qquad (8)$$

The future speed $V_j^{i+1}$ can surely indicate the traffic efficiency payoff, while the speed increment $V_j^{i+1} - V_j^i$ can represent a correct direction of improvement. Considering a recommended speed $v^{exp}$ in the intersection, for vehicle $j$ an objective indicator of traffic efficiency payoff is obtained, i.e. the speed advantage $v_j^i$

$$v_j^i = w_{vref} * V_j^{i+1} + w_v * (V_j^{i+1} - V_j^i) \qquad (9)$$

where $w_{vref} = 1/v^{exp}$, and $w_v$ are weighting coefficients, here $w_v$=0.5.

**3.3 Subjective payoff modelling for human-likeness**

It is tempted to formulate a total payoff as a weighted sum of objective indicators in Eqs. (7) and (9), e.g. $0.2A_s^i + 0.8v_j^i$ for an efficiency-oriented driving style. However, in real driving scenarios, even if the vehicle is fully automated and purely rational in decision, the subjective feelings of the onboard passengers cannot be ignored, which is actually an important indicator of user experience. Therefore, here an attempt is made to consider human subjective feelings of safety and speed, to capture real driver styles of decision in interactions.

3.3.1 Subjective payoffs of safety and speed

Prospect Theory, developed by Kahneman and Tversky, is a behavioral model of human choice, which finds broad applications in economics.[25] Since this model shows how people decide between alternatives that involve risk and uncertainty, naturally it can be borrowed for the subjective payoff description here, i.e. how the objective indicators of the strategies can be valued as an individual's subjective payoffs. Prospect Theory believes that there are individual differences in the value function, which also change with individual preferences and contexts of decision.[26] With this, three characteristics of human decision irrationality, i.e. reference dependence, saturation of gains and losses, and loss aversion, can be captured. Based on Prospect Theory, a constant interval $t_{safe}$ is chosen as the reference, and then the subjective safety payoff is

$$f(t_i) = \begin{cases} (A_s^i - t_{safe})^\alpha, & A_s^i \geq t_{safe} \\ -\lambda(A_s^i - t_{safe})^\beta, & A_s^i < t_{safe} \end{cases} \qquad (10)$$

where $\alpha, \beta, \lambda$ are model parameters, and $\alpha$=$\beta$=0.88, $\lambda$=2.25.

Considering human decision with saturation of gains and losses, the final subjective speed payoff is modelled as

$$g(v_j^i) = K\left(1 - \theta^{(v_j^i/v_{ref})}\right) \qquad (11)$$

where $v_{ref}$ is a reference speed in intersection driving, $K$ limits the payoff value, and $\theta$ determines its increasing trend. As shown in Fig. 4(b), with $K$=1.142 and $\theta$=0.26, the speed payoff increases rapidly when the vehicle speed is low and saturates at high speeds due to the exponential form.

Therefore, for the total payoff of vehicle $j$,

$$U_j^i = \sigma_j f(A_s^i) + \delta_j g(v_j^i) \qquad (12)$$

where coefficients $\sigma_j$ and $\delta_j$ indicate the driver weightings on safety and speed, respectively. Here a constraint $\sigma_j + \delta_j = 1$, $\sigma_j, \delta_j \in R$ is adopted. Note that a normal driver usually has $\sigma_j$ and $\delta_j$ in the range of (0,1), representing a tradeoff between safety and speed.

Combination of weightings $\sigma_j$ and $\delta_j$, together with $f$ and $g$, allows freedom in modelling human decision preferences. The variation of driver preferences during a decision process can be captured with corresponding changes of weightings $\sigma_j$ and $\delta_j$.

3.3.2 Acceleration tendency in interaction

Driver's acceleration tendency is the probability that the driver adopts acceleration strategy to avoid collision at intersections. Referring to the time to arrival (TTA) defined in Eq. (1), if arrival times $t_A - t_B < 0$, Car A has the priority to pass the intersection, then it will tend to accelerate, and the earlier Car A arrives than Car B, the more remarkable the acceleration tendency of Car A is. If arrival times $t_A > t_B$, Car B arrives first, and Car A tends to slow down and to yield for Car B. But when the TTA difference $t_A - t_B$ is big enough, e.g. $t_A - t_B$ is larger than a minimum safety time interval of 1.5 seconds according to the study of Meng et al.,[27] Car B reserves some safety space and is supposed to have acceleration tendency again. To sum up, the driver's acceleration tendency of vehicle $j$ is indicated by probability $p_j$ as follows,

$$p_j = \begin{cases} \max((t_{\bar{j}} - t_j)/t_{\bar{j}}, \epsilon), & t_j \leq t_{\bar{j}}; \\ \max(1 - \exp(0.5 - 0.5t_j/t_{\bar{j}}), \epsilon), & t_j - t_{\bar{j}} \geq 1.5; \\ \epsilon, & \text{other condition.} \end{cases}$$

(13)

where $\epsilon$ is a small positive number indicating a minimum acceleration probability. An example acceleration probability of vehicle $j$ is shown in Fig. 4(c), with $\epsilon = 0.05$ and the other vehicle TTA as $t_{\bar{j}} = 3$ seconds.

Therefore, by regarding the acceleration probability $p_j$ as a discount factor on the total payoff in Eq. (12), the final payoff of a strategy is predicted using

$$u_j^i = p_j * \left(\sigma_j f(A_s^i) + \delta_j g(v_j^i)\right) \quad (14)$$

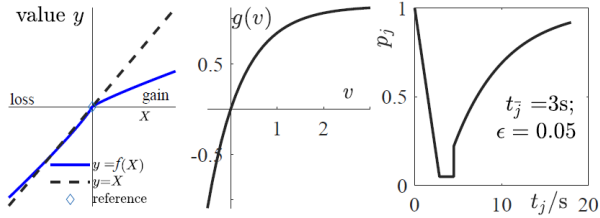

(a) subjective safety value function, (b) subjective speed payoff, (c) acceleration probability

Fig. 4 Human-likeness modelling of decision payoffs

### 3.4 Driving style identification

It is hopeful to improve the driving decision of one vehicle if the other vehicle's driving style can be identified in time. This is necessary in application of interacting with human drivers, who may vary their driving styles even in one interaction. Although the AV performances when interacting with human drivers are not within the scope of this study, the main idea is briefly described as follow. As shown in Fig. 3, here the theory of feedback is used to simplify the driving style identification, i.e. using a compensation gain $k_\sigma$. When the actual acceleration of the interaction vehicle is different from the estimated acceleration based on Nash equilibrium, a negative compensation is adopted in the direction of the actual acceleration $\hat{\sigma}_{j,\bar{j},new} = \hat{\sigma}_{j,\bar{j}} - k_\sigma * (a_j - a_j^*)$. Here, $a_j^*$ represents the optimal acceleration of the interaction vehicle at the last decision step.

### 3.5 Nash equilibrium

Given the payoff function in Eq. (14), the payoff matrix for all the possible strategies combinations can be found. Then the solution to the intersection decision problem is the best strategy satisfying the Nash equilibrium.[28] In a game $G = \{S_1, \dots, S_n: u_1, \dots, u_n\}$ with the strategy profile $(s_1, s_2 \dots, s_n)$ consisting of strategies from every game player, if the strategy $s_j$ of any player $j$ is the best countermeasure for the strategy profile of the remaining players $(1, \dots s_{j-1}, s_{j+1} \dots, s_n)$, i.e.

$$u_i^*(s_1^*, \dots s_{j-1}^*, s_j^*, s_{j+1}^*, \dots, s_n^*)$$
$$\geq u_i(s_1^*, \dots s_{j-1}^*, s_j, s_{j+1}^*, \dots, s_n^*) \quad (15)$$

that holds for any $s_j \in S_j$, then $(s_1^*, \dots s_{j-1}^*, s_j^*, s_{j+1}^*, \dots, s_n^*)$ is defined as a Nash equilibrium of the game $G$.

Note that in some cases, the 2-by-2 game may result in multiple solutions or no solution to Nash equilibrium. In these cases, the strategy can be selected by setting the natural rules as follow.
a) Case 1: if there is exactly one solution, then the two vehicles execute the strategy according to the Nash equilibrium. This indicates that the driver has made a clear and unambiguous optimal decision in the conflict situation.
b) Case 2: if there is no solution, which indicates that the two vehicles are in an unsafe state of interaction, the drivers need to pay more attention to the safety factors when crossing the intersection. The actual practice is to increase $\delta$ in Eq. (14) and then play the game again until the Nash equilibrium has a solution. This means that drivers will adopt more cautious strategies when the conflict situations become more uncertain.
c) Case 3: if there are two solutions, e.g. $(a_{Acc}, a_{Dec})$ and $(a_{Dec}, a_{Acc})$ are both the solutions satisfying the Nash equilibrium, then the same strategy as performed in the last subgame is preferred. This mimics a kind of inertia in human decision making. Generally speaking, one does not intentionally quit the current superior strategy to pursue another equally superior strategy. If these two solutions are different from the last strategy, then the optimum solution $(s_A^*, s_B^*)$ is the one with the largest total payoff, i.e.

$$(s_A^*, s_B^*) = \underset{s_A, s_B \in \{a_{Acc}, a_{Dec}\}}{\operatorname{argmax}} \sum_{i=1}^{2} u_i(s_A, s_B) \quad (16)$$

## 4. Simulation and results

### 4.1 Simulation methodology

Joint simulation of two or four AVs is carried out to validate the proposed decision framework, especially in the limit cases with various initial conditions and driving style. As depicted in the two-AV interaction example of Fig. 5, MSC Carsim®, an extensively adopted commercial software, is used to model the system dynamics of both vehicles, especially the longitudinal dynamics. The intersection scenario is also built in CarSim. The position, velocity and acceleration of both vehicles are sent to the decision module in Matlab/Simulink®, which outputs the target acceleration based on the aforementioned game model. Both AVs make decisions based on the proposed algorithm. Similarly, interactions of four AVs in a four-arm intersection are simulated in Section 4.4.

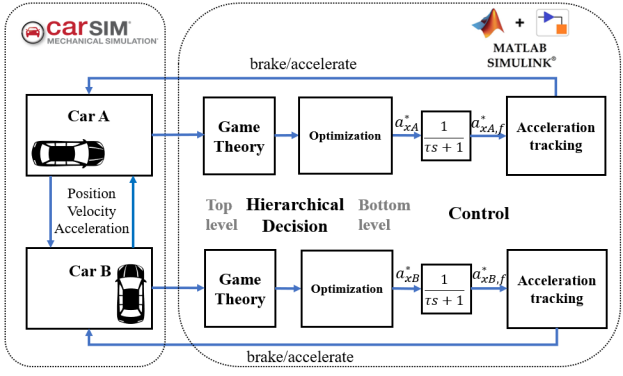

### 4.2 Limit cases with various initial speeds and distances

Considering a general first-in-first-out rule in traffic, the indistinction of right of way is apt to cause vehicle collision at intersection. To this end, a first setting of simulation condition is selected as follows: the initial speeds of Cars A and B are both 40km/h, the initial acceleration is 0, and the distance to the nearest side of the conflict area is 60m, i.e. $d_{A0} = d_{B0} = 60$. The example safety weightings of two vehicles are set as $\sigma_A = 0.6$ and $\sigma_B = 0.5$, meaning Car A is more cautious than Car B. The duration of every subgame is 0.5s, while the simulation running frequency is 1000Hz. In order to avoid the decision dilemma with exactly equal initial states of two vehicles, a random disturbance $\zeta$ with normal distribution is added to the vehicle speeds, i.e. $\zeta \sim N(0, 0.001)$ in m/s.

In this section of limit case simulation, the top level decision to accelerate or to decelerate is the focus, so the inter-vehicle interaction is thought to end when the early vehicle arrives at the conflict zone. For the rest of time in one intersection case, the late-vehicle is assumed to yield, while its bottom level decision is not detailed here. Referring to Fig. 2, the residual clearance $\Delta_d$ and the interaction duration $T_f$ are used to judge the safety and efficiency of intersection interactions. For visual comparisons in multiple cases, a unitary judgement limit $\Delta_d = 3m$ is used in the simulated cases. Note that for real applications, this limit should be calibrated with consideration of the late-vehicle's braking capabilities.

Fig. 6 shows the simulation results, including vehicle states, acceleration probability and payoffs. For payoffs, $u_{s_A, s_B}$ indicate the payoff of the two vehicles taking strategy $s_A$ and $s_B$, respectively. $s_A, s_B \in \{p = a_{Acc}, n = a_{Dec}\}$, e.g. $u_{np}$ means the payoff when Car A decelerates and Car B accelerates. In the first sub-game, from 0 to 0.5s, when both vehicles have similar speeds and distances, the best decision for both vehicles is to accelerate Car A and to decelerate Car B. In the middle stage, from 0.5s to 3s, both vehicles continue to maintain this strategy until there is big enough TTA difference (i.e. safety margin). Finally, the slower vehicle, Car B, switches from braking to acceleration, to increase its speed $V_B$. As shown in Fig. 6(b), Car B does not have to fully stop to give way, but can cross the intersection at a relatively high average speed after 3s, which improves the traffic efficiency of the intersection. As shown in Fig. 6(c), at 4.63s, when Car A reaches the intersection, the residual clearance $\Delta_d$ of Car B is 15.08m.

Fig. 6(d) shows the predicted acceleration probabilities of both vehicles by Eq. (13), which is of help to understand the decision process. In the first 1 second, there is a high conflict risk and the acceleration probabilities of both vehicles, $p_A \& p_B$, are relatively low. During 1-3.5s, $p_A \& p_B$ increase with the growing safety margin. Then at 3.5s, the two vehicles are both close to the intersection, but due to higher speed of Car A, the arrival time $t_A < t_B$, which results in a decrease in the acceleration probability of Car A. This also indicates that when both vehicles are close to the intersection, the two-vehicle interaction has a great impact on acceleration tendency and $t_B - t_A \geq 1.5s$. Even though Car B is still safe to accelerate, the driver's cautious preference in approaching the intersection will not allow acceleration, thus avoiding a collision. Therefore, it can be found that the proposed framework can capture the driver characteristics in intersection interaction.

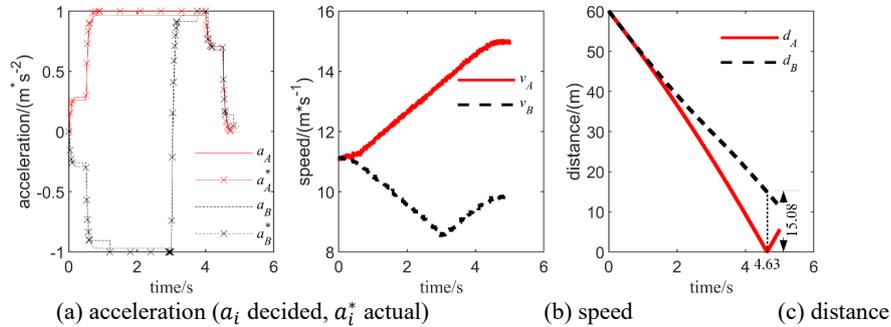

(a) acceleration ($a_i$ decided, $a_i^*$ actual)   (b) speed   (c) distance

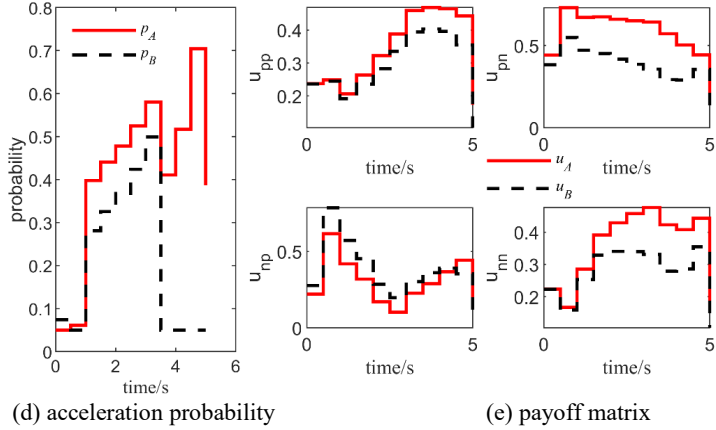

(d) acceleration probability      (e) payoff matrix

Simulations with different initial conditions at higher initial speeds and shorter initial distances are carried out, of which the results are summarized in Table 1. In cases of $d_{A0}=d_{B0}$, the decision-making of both vehicles is quite difficult with the same initial speed and acceleration. In the simulation setup 1, as the initial speed increases, the entire interaction duration $T_f$ decreases from 4.630s to 2.043s, while the residual clearance $\Delta_d$ is always above 5m. When the initial speed increases to 100 km/h, which though rarely happens in real life, $\Delta_d$ decreases to 6.136 m, and the two vehicles can still pass the intersection safely. And further reduce the initial distances $d_{A0} = d_{B0}$ to 50m, and vary the initial speed from 40 to 100 km/h. It can be found that the residual clearances $\Delta_d$ are all less than that of $d_{A0} = d_{B0} = 60$m. When the initial speed equals to or is lower than 90 km/h, both vehicles can safely interact. But when the initial speed reaches 100km/h, the whole vehicle interaction process lasts only 1.725s, and the residual clearance $\Delta_d$ is only 3.422m, which means almost collision.

Table 1 Simulation results of different vehicle initial speeds

| Setup 1 | $d_{A0} = d_{B0} = 60$m | | | | | | |
|---|---|---|---|---|---|---|---|
| $V_{A0}$ /(km/h) | 40 | 50 | 60 | 70 | 80 | 90 | 100 |
| $V_{B0}$ /(km/h) | 40 | 50 | 60 | 70 | 80 | 90 | 100 |
| $T_f$/s | 4.630 | 3.893 | 3.109 | 2.769 | 2.477 | 2.239 | 2.043 |
| $\Delta_d$/m | 15.08 | 12.89 | 15.53 | 11.63 | 9.349 | 7.596 | 6.136 |
| Setup 2 | $d_{A0} = d_{B0} = 50$m | | | | | | |
| $V_{A0}$ /(km/h) | 40 | 50 | 60 | 70 | 80 | 90 | 100 |
| $V_{B0}$ /(km/h) | 40 | 50 | 60 | 70 | 80 | 90 | 100 |
| $T_f$/s | 3.551 | 3.061 | 2.656 | 2.358 | 2.101 | 1.893 | **1.725** |
| $\Delta_d$/m | 10.39 | 10.45 | 10.76 | 7.734 | 6.168 | 5.76 | **3.422** |

To further validate the algorithm's effectiveness and its capability limits at different initial vehicle positions, multiple rounds of simulation with an initial speed of 40km/h are carried out. The initial distance $d_{B0}$ varies from 40m, 50m, ..., to 100m, while the initial distance for Car A is selected as $d_{A0} = x + d_{B0}, x \in \{-20, -19, ..., 0, ..., 19, 20\}$. Fig. 7 shows the simulation results. Given $d_{B0}$, the residual clearance $\Delta_d$ varies with $d_{A0}$ in a V-shape fashion. Unsurprisingly, the smallest residual clearances $\Delta_{d,\min}$, which are most risky, come with initial $d_{A0}$ that is close to $d_{B0}$. When the initial positions of both vehicles are close to the intersection, e.g. $d_{B0} = 40$m, the residual clearances in some limit cases are very close to, but are still larger than, the collision judgement limit, which means that both safety and traffic efficiency can be guaranteed.

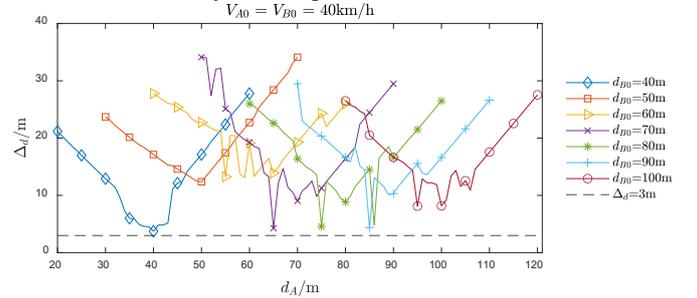

Fig. 7 Residual clearance results with initial speed 40km/h and different initial distances

### 4.3 Uniformly-sampled cases

Different from the limit cases in Section 4.2, here uniformly-sampled cases are simulated, which are more common in real traffic. The initial conditions are set as follows. The initial distance $d_{A0}$ of Car A is 40 to 80m, uniformly sampled every 1m. The initial speed $V_{A0}$ of Car A is 9 to 13m /s, uniformly sampled at 0.1m/s; correspondingly, the initial speed $V_{B0}$ of Car B is $V_{A0} - 2.5$ m/s to $V_{A0} + 2.5$ m/s, uniformly sampled every 0.1m/s. The initial distance $d_{B0}$ of Car B satisfies $|d_{B0}/V_{B0} - d_{A0}/V_{A0}| < 0.5$ s, with which the two-vehicle conflict is relatively intense due to the indistinction of the right of way. Therefore, a total of 82,000 cases of initial conditions are generated.

The decision model based on leader-follower game theory by Li et al is chosen as the benchmark algorithm.[12] They set the vehicle that does not have the right of way in the intersection as the follower, which can only make a benefit maximization decision, i.e. a maximin strategy, by considering the worst possible behavior of the interaction vehicle. The vehicle with the right of way is set as the leader, which knows that the follower's decision is the maximin strategy. The leader-follower algorithm (LF) is implemented with the original parameter setting from Ref. 12, i.e. 1s of decision interval and the acceleration between -4 and 2 $m/s^2$. For the fair comparison, the proposed algorithm (PT) is also set as so.

Statistics show that the collision rate of LF is 11.43%, which means that in such failed cases emergency braking should be activated. The collision rate of PT is only 1.90%, meaning that the proposed algorithm is safer than the benchmark algorithm. Fig. 8 details the distribution of emergency cases of different initial conditions. LF is more difficult to make correct decisions when the initial distance is larger or the initial speed is higher, but it performs slightly better than PT when the initial vehicle position is closer to the intersection. This can be explained by the differences of payoff designs for the two algorithms. For PT, it adopts TTA as a safety payoff indicator, and thus can detect and attempt to reconcile a potential collision conflict at the beginning of the two-vehicle interaction. In contrast, the LF algorithm cannot avoid a potential collision until it is predicted to happen within a prediction horizon, which is set 2s according to Ref. 10. The host vehicle will keep accelerating when the interacting vehicle is out of the prediction horizon, which leads to worse decisions at larger initial distances or higher velocities. Based on the current setting of decision parameters, the PT algorithm is suitable for unsignalized intersections with lighter traffic, e.g. those in rural or suburb areas. For heavier traffic cases with lower initial speeds and smaller initial distances, the residual interval $\Delta t^i$ in the safety payoff of the PT algorithm is too small to work as an effective indicator of safety. In such cases, the potential improvement can be made by combing $\Delta t^i$ with the current distance $d_j^i$.

As shown in Fig. 9, a specific case is analyzed and discussed for this phenomenon in detail. It can be seen that LF fails in this case, but PT successfully avoids the collision. In view of the potential collision at t=0s, the PT algorithm successfully decides a collision avoidance operation, i.e. Car A braking and Car B accelerating. If with guaranteed safety, vehicles may accelerate again to increase speed, e.g. Car A at t=1s. In contrast, the LF algorithm does not detect the potential collision at t=0s, leading to both cars' acceleration decision. At t=1s, a collision is predicted, and both cars decide to decelerate to reduce the potential collision severity. Neither cars do not start any collision avoidance operation until t=2s, when it is too late to avoid the collision. For the LF algorithm, a potential improvement of safety performance can be made via increasing the prediction horizon, but the computing load increases correspondingly. On the other hand, an increased prediction horizon will lead to earlier deceleration decisions of both cars, which works more conservative and affects the traffic efficiency.

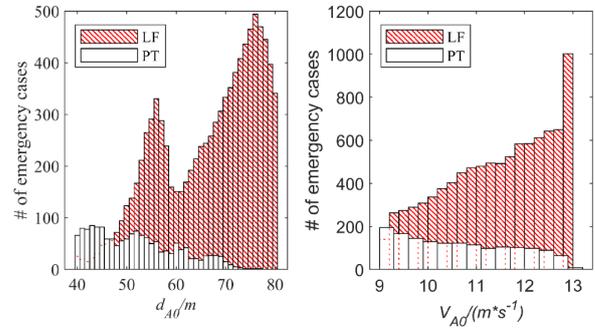

Fig. 8 Statistics of emergency cases with two algorithms

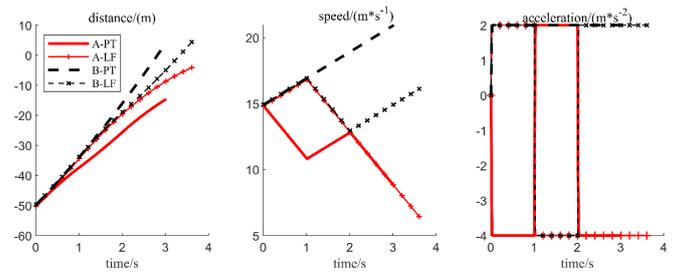

Fig. 9 Interaction process comparison in an example case study (Ts=1s)

## 4.4 Four-AV interaction cases at a four-arm intersection

As shown in Fig. 10, four-AV interaction cases at a four-arm intersection are simulated to validate the algorithm performances, especially the reasonability of reducing multi-vehicle conflicts into two-vehicle interaction sub-problems. The complexities of four-AV interaction are obviously increased by adding two more players in the game, while each player may potentially conflict with two other players in two conflict areas. A global coordinate system xoy is defined as shown in Fig. 10. The car position is defined using TTS, i.e. its time to stopline, while the positive or negative sign of TTS indicates whether the car lies before or after the stopline.

One extreme initial condition is set as follow: the initial TTS is 6s, the initial velocity is uniformly distributed as $V_0 \sim U(10,14)$ (m/s), and the initial acceleration is uniformly distributed as $a_0 \sim U(0,4)$ m/s². Therefore, for the randomly generated case to be discussed, the initial velocities for Cars A~D are 11.87, 13.58, 12.54 and 10.50 m/s, respectively, while their initial accelerations are 3.68, 2.35, 2.15 and 0.33 m/s², respectively.

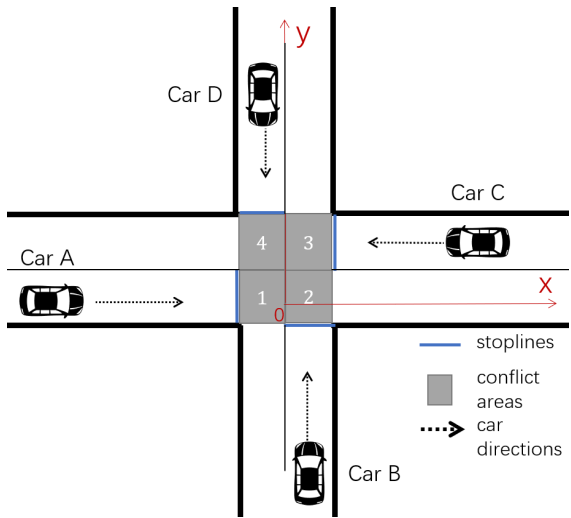

Fig. 10 Four-AV interaction scenario in a four-arm intersection

The results of this interaction case are summarized in Fig. 11, including locations and motion states during the entire interactions. Since the initial TTS is large as 6s, all cars decide to accelerate during the initial 1.5s. Car A has the right of way and thus applies the maximum acceleration. It is closest to the stopline at t=1.5s, with TTS only 3.4s, so Cars B and D can only decelerate to avoid conflict with Car A. On the other hand, Cars B and D also interact with Car C, and the game solutions for B and D suggest them to accelerate, since Car C does not have any advantage of the right of way and is to decelerate from t=1.5s. According to the safety-first principle in Eq. 2 in Section 2.1, Cars B and D should decelerate. Based on the proposed multi-vehicle game simplification approach, Car C can not consider the influence of Car A on Cars B and D, although Car C may accelerate by potentially taking advange of Car A's cover activity. Such loss of time efficiency may be viewed as a side effect of the multi-vehicle game simplification, however, it is reasonable to put safety first in multi-vehicle interactions. At t=3s, Car D has already distanced far enough away from Car A, so it should accelerate. In its simutaneous interaction with Car C, Car D also has advantage of the right of way and should accelerate. Therefore, Car D starts to accelerate from t=3s, then stops acceleration at t=5.5s to maintain its speed and distance, and finally crosses the intersection at t=6.21s.

Fig. 11(a) also presents the arriving and leaving time of four cars in four conflict areas, showing that the sequence of passing the intersection is A, D, C and finally B. To quantify the time efficiency of such four-car interactions, the intersection clearing time $T_{IC}$ is used, which is defined as the time that the slowest car uses to pass the intersection. Note that in uncontrolled case, if all cars maintain the initial velocities in crossing the intersection, the sequence of passing is A, B, C and then D. Although such uncontrolled case means no active interactions happen, and there may be collisions between cars, it can still work as a time efficiency benchmark. For this example case, the intersection clearing time is $T_{IC} = 7.49s$, i.e. the passing time of Car B, while the benchmark clearing time is the passing time of Car D, i.e. $T_{IC0} = 7.33s$. It can be found that the fastest Car A at initial time is affected least, since it keeps the advantageous right of way all through the interactions. It is interesting that the sequence of passing is changed, i.e. the slowest Car D at initial time can actually cross the intersection earlier than Cars B and C. This shows that the increased dynamics in multi-vehicle interaction can significantly affect the players's decision behaviors. With the proposed approach of simplification, time efficiency may be compromised in extreme conditions, but interaction safety can still be guranteed.

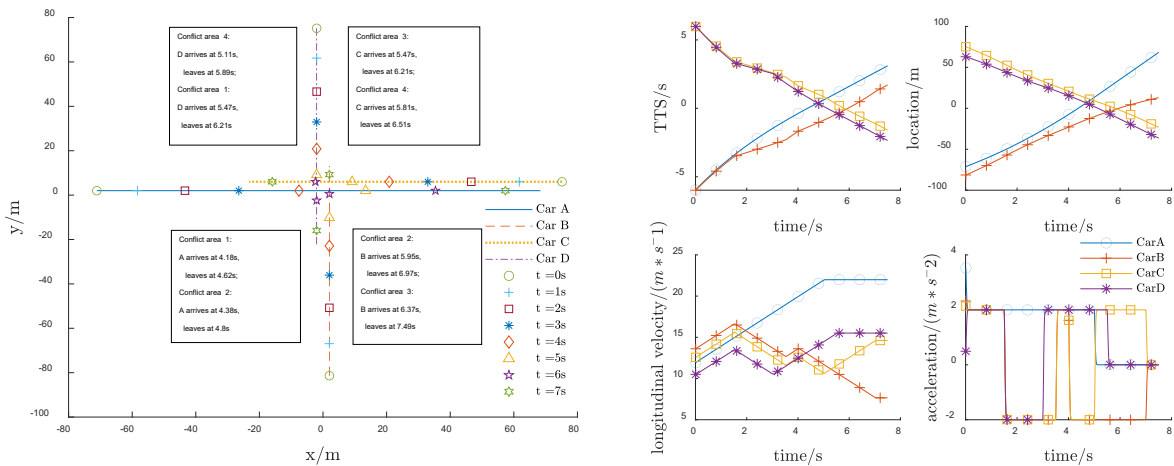

(a) Car locations at different time points; (b) Car positions, velocities and accelerations

Fig. 11 Results of one extreme four-AV interaction case (Ts=0.5s)

The interaction scenario in Fig. 10 is further expanded by the random sampling of initial car locations, while the velocity and acceleration are set as $V_0 \sim U(10,14)$ and $a_0 \sim U(0,4)$, respectively. Totally 9 groups of cases are simulated with different levels of interaction conflict, i.e. the initial TTS for each car is chosen as $TTS = 6 + \mu * U(0,1)$ s, where $\mu=0, 1, \ldots, 8$ is the bandwidth of TTS uncertainty. 10,000 simulations are carried out for each case group, while both the success rate (SuccessRate) and time efficiency are discussed. For time efficiency, the mean value of intersection clearing time $T_{mIC}$ is used to compare with the expectation value of benchmark clearing time $T_{eIC0}$, which can be derived as $T_{eIC0} = 7.2037 + 0.8\mu$ in the simulation. Therefore, the potential time efficiency improvement percentage is defined as $\eta_{TE} = (T_{eIC0} - T_{mIC})/T_{eIC0} * 100\%$.

The results of success rate and time efficiency are summarized in Table. 2. With the smallest initial car distances, i.e. $\mu = 0$, the success rate of interaction is 86.94%, while the time efficiency is only improved by 2.54%. With increasing $\mu$, the level of potential interaction conflict decreases, both the success rate and time efficiency improvement increase. In case group 9, $\mu = 8$, the initial time to arrive at the intersection of four cars may differ much, the success rate reaches 96% and the time efficiency can be improved by 34%. This is reasonable in that if collision conflict is more likely to happen, the AV driver tends to be more conservative and leads to lower time efficiency.

Table 2 Statistical results of sampled four-AV interaction cases

| Case group's $\mu$ value | SuccessRate (%) | $T_{mIC}$ (s) | $T_{eIC0}$ (s) | $\eta_{TE}$ (%) |
|---|---|---|---|---|
| 0 | 86.94 | 7.0204 | 7.2037 | 2.54 |
| 1 | 89.14 | 7.4016 | 8.0037 | 7.52 |
| 2 | 89.32 | 7.5921 | 8.8037 | 13.76 |
| 3 | 90.72 | 7.7432 | 9.6037 | 19.37 |
| 4 | 92.7 | 7.9094 | 10.4037 | 23.98 |
| 5 | 93.9 | 8.1429 | 11.2037 | 27.32 |
| 6 | 94.71 | 8.3786 | 12.0037 | 30.2 |
| 7 | 95.83 | 8.6686 | 12.8037 | 32.3 |
| 8 | 96.08 | 8.9776 | 13.6037 | 34.01 |

The above results of simulation show that our algorithm can perform appropriate interaction decisions at unsignalized intersections. Thanks to the human-like payoff functions, the proposed game-theoretic framework can capture the dynamic interaction characteristics under various initial conditions of intersection driving, by optimizing the tradeoff between safety and speed. The proposed approach can also reasonably describe the multi-vehicle interaction process, while guaranteeing safety and improving time efficiency.

## 5. Conclusion

Unsignalized intersection driving is challenging for both human drivers and automated vehicles. In this paper, the problem is formulated as a multi-stage repeated game, and a methodology of human-like payoff design is proposed to consider human driver subjective feelings and dynamic tradeoff of safety and traffic efficiency. The effectiveness of the game-based decision algorithm is validated via simulation.

This work is a part of an on-going research on human-like automated driving, which aims to develop flexible and socially-aware decision and planning algorithms. In future work, it is necessary to carry out more extensive simulations, especially with driving simulator and real vehicle experiments, to validate the decision performances of the proposed algorithm with human drivers. The human-likeness consideration should be further quantified and justified with human evaluations. Other directions of research may include verification and validation for the more complicated multi-vehicle conflict situations.

## Supplementary materials

A supplementary video abstract is provided, which includes a brief introduction of the proposed approach. Please check: https://space.bilibili.com/544424063 .

## Acknowledgements

The authors appreciated Linhui Chen for his help in preparing the video abstract.

## Declaration of conflicting interests

The author(s) declared no potential conflicts of interest with respect to the research, authorship, and/or publication of this article.

## Funding

The author(s) disclosed receipt of the following financial support for the research, authorship, and/or publication of this article: This work was supported by Department of Science and Technology of Zhejiang (No. 2021C01SA601840, 2018C01058).

## ORCID iDs

Daofei Li, https://orcid.org/0000-0002-6909-0169
Guanming Liu, https://orcid.org/0000-0002-2073-0460
Bin Xiao, https://orcid.org/0000-0001-9752-7842

## Nomenclatures

| NOTATION | VARIABLE | NOTATION | VARIABLE |
|---|---|---|---|
| $a_{Acc}$ | Acceleration strategy | $T_f$ | Entire interaction duration |
| $a_{Dec}$ | Deceleration strategy | $T_{IC}$ | Intersection clearing time |
| $a_{expj}$ | Expected acceleration strategy of vehicle $j$ | $T_{mIC}$ | Mean value of intersection clearing time |
| $a_j^i$ | Acceleration of vehicle $j$ | TTS | Time to stopline |
| $a_x^*$ | Target acceleration | $U_j^i$ | Total payoff of vehicle $j$ |
| $A_s^i$ | Safety advantage | $v^{exp}$ | Recommended speed |
| $d_j^i$ | Distance from the head of vehicle $j$ to the near edge of the conflict area | $v_j^i$ | Expected speed |
| $d_{j0}$ | Initial distance from the head of vehicle $j$ to the near edge of the conflict area | $V_j^i$ | Current speed of vehicle $j$ |
| $D_j^i$ | Distance from the tail of vehicle $j$ to the far edge of the conflict area | $V_{j0}^i$ | Initial speed of vehicle $j$ |
| $f(\cdot)$ | Subjective safety payoff function for driver | $\alpha, \beta$ | Coefficients of the gains and losses in Prospect Theory |
| $g(\cdot)$ | Subjective speed payoff function for driver | $\delta_j$ | Speed weighting of vehicle $j$ |
| $i$ | $i$-th subgame, used in superscript | $\epsilon$ | Small positive number |
| $j$ | Index for vehicle A or B, used in subscript | $\Delta_d$ | Residual clearance defined in Fig 2, $\Delta_{d,\min}$ Is the smallest residual clearance |
| $K$ | Maximum subjective speed payoff in Eq. (11) | $\Delta t^i$ | Estimated residual interval |
| $p_j$ | Acceleration probability of driver in vehicle $j$ | $\Delta t_{exp}^i$ | Expected residual interval |
| $s_j^*$ | Optimum strategy taken by vehicle $j$ | $\eta_{TE}$ | Time efficiency improvement percentage |
| SuccessRate | Success rate of multi-vehicle interation | $\theta$ | Parameter of subjective speed payoff |
| $t_j^i$ | Time to arrive (TTA) of vehicle $j$, the time that it needs to arrive at the conflict area | $\lambda$ | Loss aversion coefficient in Prospect Theory |
| $t_{max}$ | Maximum value of $t_j^i$ | $\sigma_j$ | Safety weighting of vehicle $j$ |
| $T$ | Subgame duration | $\tau$ | First order delay constant of acceleration actuator |
| $T_{eIC0}$ | Benchmark intersection clearing time | $\tau_j^i$ | Passing time of vehicle $j$, the time it needs to completely pass the conflict area |